\documentclass[arxiv]{melba}
\usepackage{subcaption}
\usepackage{natbib}
\usepackage{mwe} 
\usepackage{tabularx}
\usepackage{array}
\usepackage{hyperref}
\usepackage{amsmath,amsfonts}
\usepackage{graphicx}
\usepackage{svg} 
\usepackage{booktabs}
\usepackage{verbatim} 
\usepackage{hyperref}
\usepackage{float}

\usepackage{tabularx}
\usepackage{booktabs}
\usepackage{array}
\usepackage{fontspec}
\usepackage{graphicx}
\usepackage{booktabs}
\usepackage{amssymb}
\usepackage{pifont}

\newcommand{\xmark}{\ding{55}}

\newfontfamily\urdufont[
    Path=./,
    UprightFont={Jameel_Noori_Nastaleeq_Regular.ttf}
]{JameelNoori}

\newcolumntype{Y}{>{\raggedright\arraybackslash}X}

\melbaid{YYYY:NNN}  
\doi{10.59275/j.melba.2024-AAAA}
\melbaauthors{X} 
\email{waqas.sultani@itu.edu.pk}
\volume{3}
\firstpageno{1}  
\melbayear{2026}  
\datesubmitted{yyyy-m1-d1}  
\datepublished{yyyy-m2-d2}  

\melbaspecialissue{Medical Imaging with Deep Learning (MIDL) 2020}
\melbaspecialissueeditors{Marleen de Bruijne, Tal Arbel, Ismail Ben Ayed, Hervé Lombaert}

\ShortHeadings{WBCMor-VQA Open Dataset}{WBCMor-VQA Open Dataset}

\title{Multilingual Hematology Visual Question Answering  Dataset}

\author{
	\name Hajra   Malik\aff{1}\orcid{0000-1111-2222-3333},
	\name Hafiza Tooba Aftab\aff{2}\orcid{0009-0007-9926-6085},
    \name Abdul Rehman\aff{1}\orcid{0000-0002-5745-7672},
    \name Mohsen Ali\aff{1}\orcid{0000-0003-4809-8679},
    \name Waqas Sultani\aff{1}\orcid{0000-0002-9322-0728}
}
\affiliations{
	\num 1 \addr Information Technology University, Lahore  \\
	\num 2 \addr King Edward Medical University, Lahore
}

\abstract{
Vision Language Models (VLMs) have shown promising capabilities in medical image analysis by jointly understanding visual and textual information for tasks such as Visual Question Answering (VQA). However, existing hematology vision-language resources remain predominantly English-centric, limiting their applicability in multilingual healthcare environments. This challenge is releveant generally to South Asia and specifically to Pakistan, where Urdu is widely used despite healthcare information and digital medical systems being largely dependent on English. To investigate this gap, we conducted a survey among healthcare professionals, which revealed substantial language mismatches between clinical documentation and patient communication, emphasizing the need for multilingual healthcare technologies. To address this limitation, we introduce \textbf{WBCMor-VQA}, a clinically validated bilingual (English--Urdu) morphology-aware VQA benchmark for leukemia and normal white blood cell (WBC) analysis. The benchmark is constructed using morphology-aware annotations from LeukemiaAttri and WBCAtt datasets and supported by a domain-specific Urdu hematology dictionary to ensure linguistic consistency and clinical correctness. The final benchmark contains 110K bilingual question–answer pairs serving as VQA annotations for 20K leukemic and normal single-cell images. Furthermore, we establish baseline performance by evaluating multiple open-source VLMs on the proposed benchmark. The proposed resource aims to facilitate the development of accessible and clinically relevant AI systems for multilingual healthcare environments. The Dataset is publicly available: https://doi.org/10.6084/m9.figshare.32727159 }


\begin{document}

\twocolumn[\maketitle]

\section{Background} 
\lettrine{L}{eukemia}, a blood cancer affecting the production and function of blood cells, is commonly diagnosed and assessed through the examination of White Blood Cells (WBCs). The disease poses a serious global health challenge because it directly impairs the human immune system, compromising its ability to fight infections \citep{bloodcancer2025}. Long-term epidemiological evidence indicates that leukemia remains one of the most prevalent hematological malignancies across diverse populations, highlighting its continued clinical and public health significance \citep{badar2023cancer}. Despite advances in diagnostic approaches and treatment strategies, leukemia continues to contribute substantially to global cancer-related morbidity and mortality. Its aggressive nature is reflected in its disproportionate impact on children and young adults, where it remains among the leading causes of cancer-related deaths in individuals younger than 45 years of age \citep{campbell2008ethnologue}.\\
In Pakistan, hematological malignancies, including leukemia, represent a considerable portion of the national cancer burden \citep{tufail2023cancer}. Although healthcare services have improved over time, cancer-related mortality remains high due to challenges such as delayed diagnosis, limited public awareness, and restricted access to specialized medical care. Recent estimates report an age-standardized cancer mortality rate of 153.52 deaths per 100,000 individuals \citep{ecancer2026}. Among pediatric populations, leukemia is of particular concern, as acute lymphoblastic leukemia (ALL) continues to be one of the most frequently diagnosed childhood cancers in the country \citep{tufail2023cancer}. Beyond the clinical challenges associated with leukemia, accessibility to healthcare information and emerging medical technologies is strongly influenced by language. South Asia, with a population of approximately 2.11 billion people, is characterized by substantial linguistic diversity. Among the major regional languages, Urdu is spoken by nearly 246 million people and serves as an important medium of communication across communities and national boundaries \citep{Poria2025}. As the national language of Pakistan and a widely understood language across South Asia, Urdu plays a critical role in facilitating access to healthcare information and improving communication between medical systems and local populations \citep{campbell2008ethnologue}.



\begin{figure*}[htb]
\centering
\includegraphics[width=0.95\textwidth]{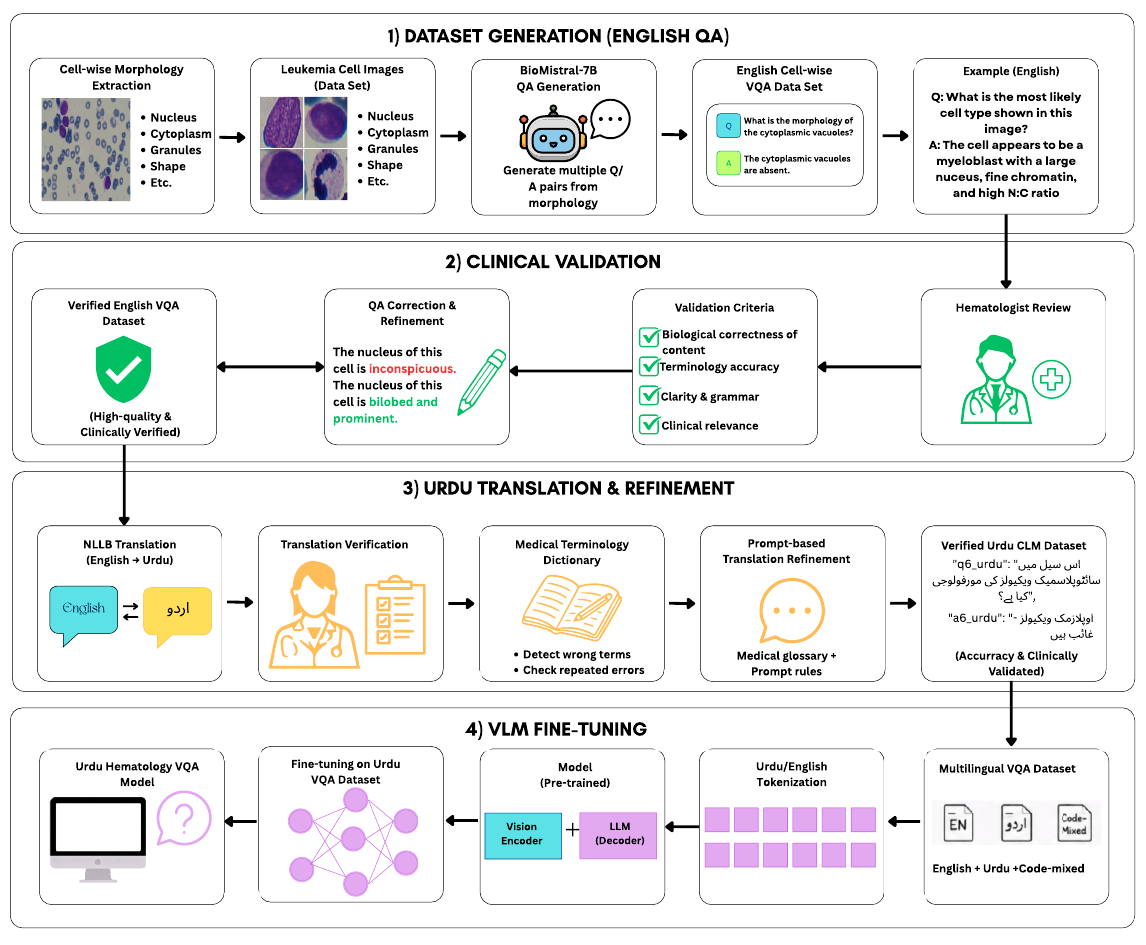}
\caption{Pipeline of the proposed bilingual WBC-VQA dataset construction and model development pipeline. Cell-level morphology attributes are converted into English VQA pairs using BioMistral-7B \citep{labrak2024}, followed by expert validation. The dataset is then translated into Urdu using NLLB \citep{nllb2022} and refined using a domain-specific medical dictionary and expert review. Finally, the resulting bilingual dataset is used to fine-tune vision-language models for hematology question answering.}
\label{fig:overall_pipeline}
\end{figure*}
Despite the widespread use of Urdu, healthcare information, medical records, and digital health technologies in Pakistan remain primarily dependent on English. Differences in literacy and English proficiency across regions and socioeconomic groups create barriers to understanding medical information and interacting effectively with healthcare systems. The evidence from the Education First (EF) English Proficiency Index 2025 rankings places Bangladesh, Pakistan, and India at 62nd, 67th, and 74th globally, suggesting that English-based healthcare and AI technologies may remain inaccessible to a large portion of the population \citep{de2025ef}. To further examine the role of language in clinical communication, we conducted a survey involving healthcare professionals who regularly interact with patients and their families. The results revealed a noticeable gap between the language used in medical documentation and the language commonly used during patient interactions. A majority of physicians (58.6\%) reported communicating with patients using a mixture of Urdu, English, and local languages, while 34.9\% primarily relied on Urdu or other local languages. In contrast, only 6.5\% reported primarily communicating in English.Additionally, 73.5\% of respondents indicated that patients often or sometimes misunderstand medical explanations, reflecting persistent communication challenges in routine clinical settings. These findings underscore the need for multilingual healthcare resources and language-adaptive medical technologies to improve communication between healthcare providers and patients. A detailed description of the survey methodology, participant demographics, and complete findings is presented in Section~\ref{sec:survey}.

Recent advances in Vision-Language Models (VLMs) have enabled artificial intelligence (AI) systems to jointly understand medical images and natural language \citep{kalpelbe2025vision, hartsock2024vision}. This capability has accelerated the development of multimodal healthcare applications, including medical image interpretation, automated report generation, and Visual Question Answering (VQA) systems \citep{bazi2023vision}. Among these applications, VQA has gained significant attention because it enables users to interact with medical images through natural language queries while receiving clinically relevant responses \citep{noshin2026survey}. Such systems offer the potential to improve access to medical knowledge, strengthen clinical decision support, and provide more interpretable AI-assisted healthcare solutions.

Building upon these developments, Uni-Hema \citep{rehman2026uni} introduced a comprehensive vision-language framework for digital hematology. The framework utilized morphology annotations from the WBCAtt dataset \citep{tsutsui2023wbcatt}, which contains normal white blood cells with detailed morphological attribute labels, to generate morphology-oriented VQA samples. In addition, leukemia-specific language supervision was incorporated through the Mask Language Modeling, enabling joint learning of visual and textual representations. More recently, HemBLIP \citep{hemblip2026} further demonstrated the effectiveness of VLM approaches for leukemia analysis by generating morphology-aware descriptions directly from microscopic blood cell images. Although these studies demonstrate the growing potential of vision-language methods in hematological analysis, existing resources remain predominantly English-centric and are primarily designed for caption generation, representation learning, or automatically generated language supervision. As a result, clinically validated multilingual resources, particularly for Urdu-speaking healthcare environments, remain largely unavailable.\\
Motivated by this limitation, we develop a clinically validated bilingual VQA resource for understanding leukemic and normal cell morphology. 


\begin{figure*}[ht]
    \centering
    
    \begin{subfigure}[b]{0.32\textwidth}
        \centering
        \includegraphics[width=\linewidth]{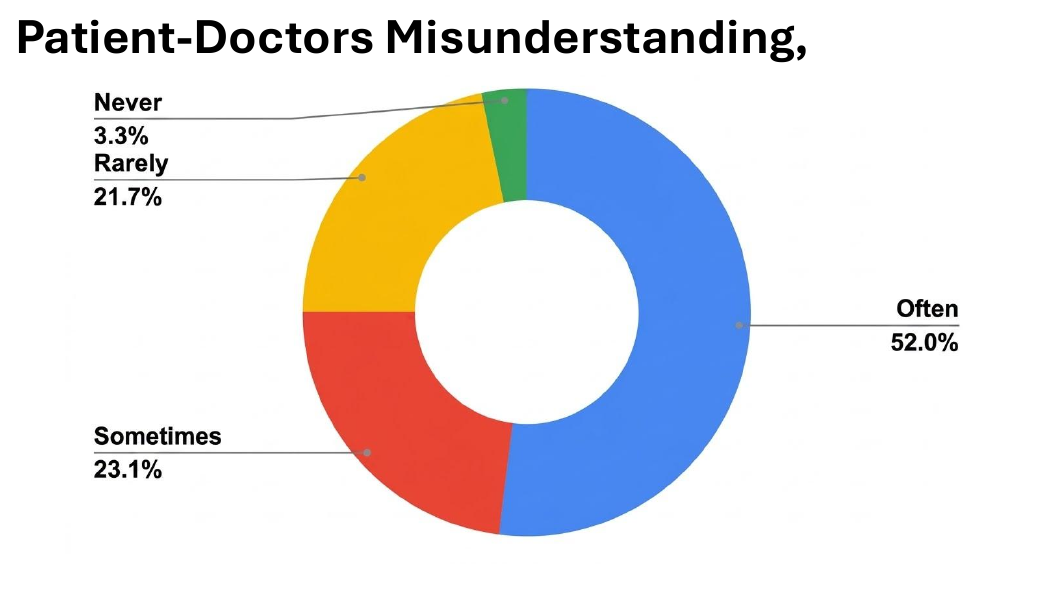}
        \caption{}
        \label{fig:misunderstanding}
    \end{subfigure}
    \hfill
    \begin{subfigure}[b]{0.32\textwidth}
        \centering
        \includegraphics[width=\linewidth]{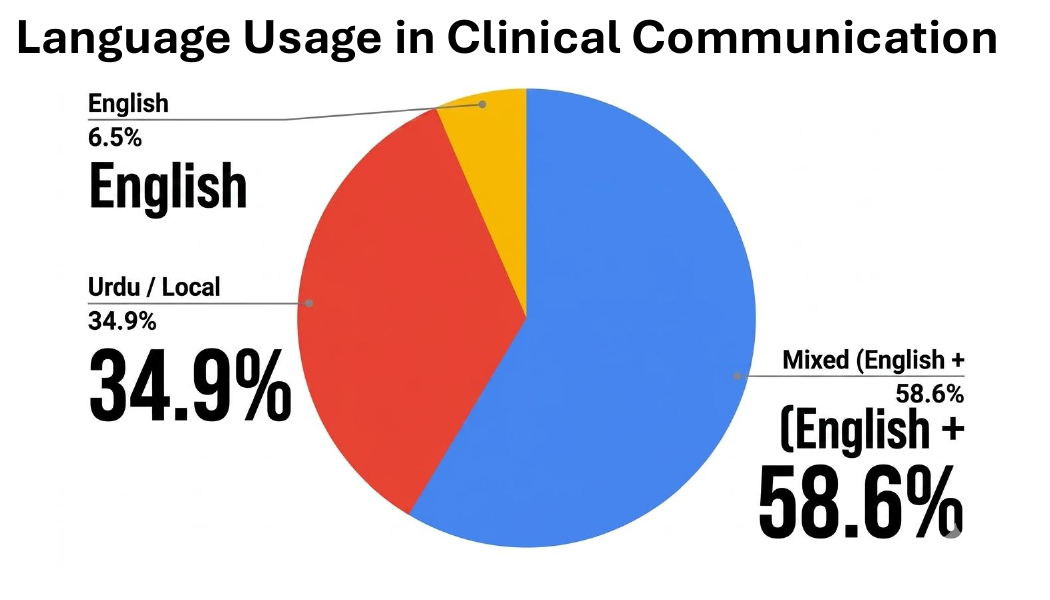}
        \caption{}
        \label{fig:language}
    \end{subfigure}
    \hfill
    \begin{subfigure}[b]{0.32\textwidth}
        \centering
        \includegraphics[width=\linewidth]{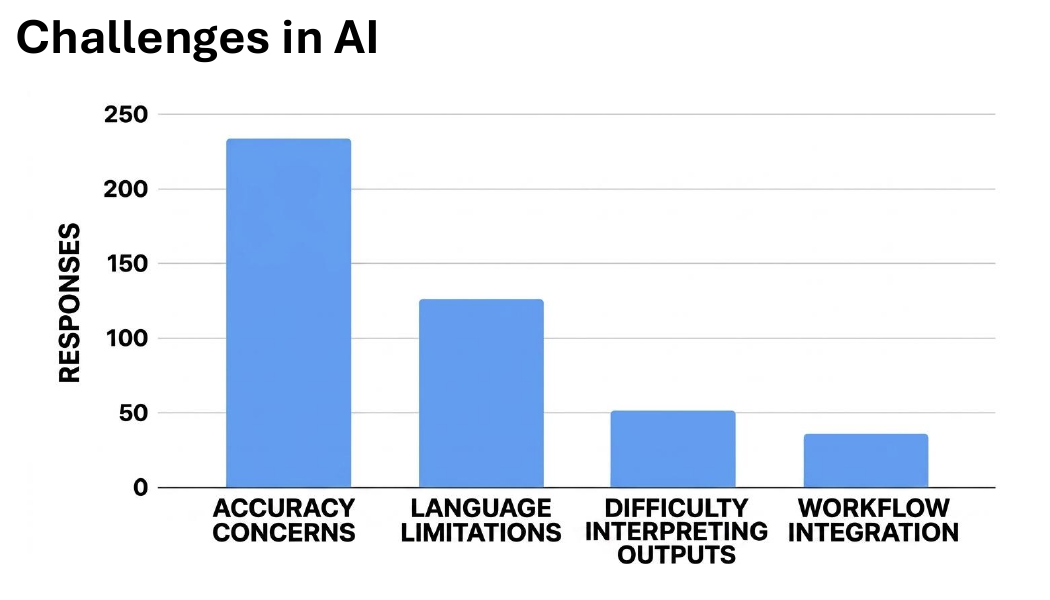}
        \caption{}
        \label{fig:ai_challenges}
    \end{subfigure}

    \caption{Survey findings related to physician communication practices and challenges in AI-assisted healthcare systems. (a) Frequency of patient misunderstanding of medical explanations, (b) language preferences used during clinical communication, and (c) challenges encountered while using AI-assisted medical systems.}
    
    \label{fig:combined}
\end{figure*}
\section{Summary}

The proposed \textbf{WBCMor\_VQA} dataset has been developed using morphology-aware annotations from the LeukemiaAttri and WBCAtt datasets \citep{rehman2024leukemiaattri, tsutsui2023wbcatt}. These datasets provide detailed morphological information for leukemia and normal white blood cell images, enabling the generation of clinically relevant visual question-answer (VQA) pairs. For the dataset curation,  we first extracted the single-cell images from the complete field-of-view microscopy images and then, following the morphology-oriented VQA generation strategy introduced in Uni-Hema \citep{rehman2026uni},  constructed English question–answer pairs and then translated them into Urdu to facilitate multilingual evaluation. To ensure both clinical validity and linguistic consistency, all generated and translated question-answer pairs were reviewed and validated by a clinical expert. The resulting \textbf{WBCMor\_VQA} resource contains paired microscopic single-cell images with bilingual annotations in English and Urdu. By integrating morphology-aware visual information with multilingual question-answer representations, the dataset supports the development and evaluation of vision-language models for leukemic and normal white blood cell morphology understanding, and broader multilingual medical AI applications.

\textbf{Our main contributions are as follows:}


\begin{itemize}
\item A cross-sectional survey involving physicians from public and private healthcare institutions was conducted to investigate language-related challenges in clinical communication and the adoption of AI-assisted medical tools. The survey explored language preferences, communication barriers, AI usage patterns, and clinicians' perspectives on improving patient understanding through AI-supported solutions.

\item We develop and clinically validate cell-level bilingual (English and Urdu) Visual Question Answering (VQA) datasets for leukemia and normal white blood cells (WBCs), comprising 110K question--answer pairs corresponding to single-cell images and enriched with morphological knowledge.

\item We construct a domain-specific Urdu hematology dictionary to standardize morphology-related terminology and ensure translation consistency across hematology question-answer pairs.

\item We evaluate multiple open-source Vision Language Models (VLMs) on the proposed bilingual datasets and provide a comprehensive analysis of their performance for hematology visual question answering in both English and Urdu.
\end{itemize}

\section{Discussion}

Effective medical communication remains a challenge in multilingual clinical environments, particularly in low and middle-income countries such as Pakistan. To better understand these challenges and the growing role of artificial intelligence in healthcare, we conducted a survey involving practicing physicians across Pakistan. The findings provide empirical evidence of communication gaps in clinical practice and motivate the need for multilingual vision-language resources, supporting the development of our bilingual hematology VQA dataset.

\subsection{Clinical Insights and Dataset Motivation}
\label{sec:survey}

A cross-sectional survey was conducted to investigate communication practices, language barriers, and the adoption of AI-assisted tools in clinical settings. The survey involved 430 practicing physicians from public and private healthcare institutions across multiple regions of Pakistan. We have made survey questionnaire available online 
\footnote{Link of the \href{https://forms.gle/4JePeDhr4RBTFeMD7}{survey form: Clinical Language and AI Usage}}.
The findings reveal substantial communication challenges in routine clinical practice, with approximately 75.1\% of respondents reporting that patients frequently or occasionally misunderstand medical explanations. To improve understanding, physicians commonly relied on mixed Urdu--English communication and simplified explanations, reflecting the multilingual nature of clinical environments and the need for bilingual healthcare resources.

The survey also highlighted the growing adoption of AI-assisted technologies in healthcare, with 73.2\% of respondents reporting the use of AI tools for tasks such as diagnostic support, report summarization, clinical documentation, and medical education. Despite this increasing integration, physicians consistently emphasized that AI systems should complement rather than replace clinical expertise, underscoring the importance of human oversight in medical decision-making. Furthermore, qualitative responses revealed a lack of standardized Urdu terminology for specialized medical domains, particularly pathology and hematology. Many physicians reported relying on English terminology or informal translations when communicating complex diagnostic concepts to patients. 


\begin{table*}[t]
\centering
\caption{Comparison of hematology morphology and vision-language datasets}
\label{tab:dataset_comparison}
\resizebox{\textwidth}{!}
{
\begin{tabular}{lcccccccc}
\toprule
Dataset &
Normal WBC &
Leukemia WBC &
Task Type &
VQA &
Bilingual &
Expert Validated &
Cell Images &
QA Pairs \\
\midrule

WBCAtt &
\checkmark &
\xmark &
Morphology Attributes &
\xmark &
\xmark &
\checkmark &
10,298 &
-- \\

LeukemiaAttri &
\xmark &
\checkmark &
Morphology Attributes &
\xmark &
\xmark &
\checkmark &
$\sim$10,000 &
-- \\

WBCAtt-VQA (Uni-Hema) &
\checkmark &
\xmark &
VQA &
\checkmark &
\xmark &
\checkmark &
10,298 & 20,944 \\

LeukemiaAttri-MLM (Uni-Hema) &
\xmark &
\checkmark &
Captioning/MLM &
\xmark &
\xmark &
\checkmark &
$\sim$10,000 &
 - \\

HemBLIP &
\checkmark &
\checkmark &
Image--Text &
\xmark &
\xmark &
Automatic &
14,659 &
- \\

\textbf{Ours} &
\textbf{\checkmark} &
\textbf{\checkmark} &
\textbf{Bilingual VQA} &
\textbf{\checkmark} &
\textbf{\checkmark} &
\textbf{\checkmark} &
\textbf{$\sim$20,000} &
\textbf{110,720} \\

\bottomrule
\end{tabular}
}
\end{table*}
The distribution of physician responses to the question shown in Figure \ref{fig:misunderstanding}, ``Have you encountered situations where patients misunderstood medical explanations?'', indicates that communication barriers are common in routine clinical practice. The majority of respondents reported that misunderstandings occur often (52.0\%) or sometimes (23.1\%), while 21.7\% reported such situations rarely and only 3.3\% reported never encountering them. Figure \ref{fig:language} presents the responses to the question, ``What language do you primarily use to communicate with patients?'' Most physicians (58.6\%) reported using a mixed English--Urdu--local language communication style, followed by Urdu or local languages (34.9\%), whereas only 6.5\% relied exclusively on English. These findings emphasize the multilingual nature of clinical communication in Pakistan. The challenges encountered by clinicians when using AI-assisted medical systems are illustrated in Figure \ref{fig:ai_challenges}. Accuracy concerns were reported most frequently, followed by workflow integration issues, difficulty interpreting AI-generated outputs, and language-related limitations. These observations suggest that concerns regarding reliability, interpretability, and integration continue to influence physician trust and adoption of AI-assisted systems. 
Collectively, these findings directly informed our dataset design. Multilingual communication motivated the inclusion of bilingual VQA pairs, while concerns about reliability and interpretability led to expert-validated annotations and medically grounded reasoning. Together, these considerations shaped the proposed English–Urdu hematology VQA dataset to support clinically relevant and linguistically accessible AI systems for medical education, decision support, and patient communication

\noindent \textbf{Limitations and Future Implicationsm :}
While the survey provides valuable insights into clinical communication practices, it is limited to physicians practicing in Pakistan and therefore may not fully generalize to healthcare systems in other regions. In addition, the findings are based on self-reported responses and may be influenced by response and recall bias. 

\section{Resource Availability}
The dataset is publicly available on Figshare \footnote{Figshare: https://doi.org/10.6084/m9.figshare.32727159}. We also provide a demo on the Hugging Face platform \footnote{Hugging Face: https://huggingface.co/spaces/ryhm/WBC-Mor-VQA}. The source code and pretrained models, along with documentation for fine-tuning, are available on GitHub \footnote{GitHub:https://github.com/intelligentMachines-ITU/WBC-Mor-VQA-dataset }.

\noindent \textbf{Potential Use Cases:} The published data can be used to train vision and large language models for research purposes only. Currently, we strongly discourage the use of the resulting models (or any model trained on given dataset) in real clinical practice or medical decision-making.

\noindent \textbf{Licensing: } The dataset, code, and pretrained models are released under the Creative Commons Attribution - NonCommercial 4.0 (CC BY-NC 4.0) license, permitting non-commercial use, sharing, and adaptation with appropriate attribution, while restricting commercial and industrial applications. All sources are properly cited.

\noindent \textbf{Ethical Considerations:} This work utilizes publicly available, clinically validated medical datasets and contains no patient-identifiable information. Following question-answer generation from the available annotations, the dataset was reviewed and validated by a medical professional. After translation, the dataset was further evaluated by two qualified native Urdu speakers, with the assistance of an English-Urdu medical dictionary, to ensure both clinical accuracy and linguistic quality.

\begin{table*}[htb]
\centering
\caption{The table shows the total number of corrections made in each category together with representative examples. During validation, experts refined unclear questions, corrected incorrect answers, fixed grammar and formatting errors, and revised medically inaccurate morphology descriptions.}
\label{tab:english_validation_5col}

\small
\begin{tabularx}{\textwidth}{lccXX}
\hline
\textbf{Correction Category} &
\textbf{Train Set} &
\textbf{Test Set} &
\textbf{Model Generated} &
\textbf{Expert Validated} \\
\hline

Question  &
09 &
84 &
What is the nuclear chromatin pattern of this \textcolor{red}{leukemia} cell? &
What is the nuclear chromatin pattern of this cell? \\

Answer  &
1238 &
761 &
The nuclear shape of this cell is \textcolor{red}{irregular}. &
The nuclear shape of this cell is \textcolor{green}{round}. \\

Grammar  &
38 &
285 &
The cytoplasm in \textcolor{red}{thisneutrophilic} cell is abundant. &
The cytoplasm in \textcolor{green}{this neutrophilic} cell is abundant. \\

Medical  &
1209 &
560 &
The nucleus of this cell is \textcolor{red}{inconspicuous}. &
The nucleus of this cell is \textcolor{green}{bilobed and prominent}. \\
\hline
Total Corrections &
1247 &
845 &
-- &
-- \\

\hline
\end{tabularx}
\end{table*}
 
\section{Methodology}
This section presents the pipeline used to construct the proposed bilingual hematology VQA benchmark. We begin with a clinical survey conducted to identify communication challenges and language requirements in healthcare settings. Next, we describe the construction of morphology-aware VQA datasets using LeukemiaAttri \citep{rehman2024leukemiaattri} and WBCAtt \citep{tsutsui2023wbcatt}, followed by expert validation, Urdu translation, translation quality assessment, terminology refinement, and prompt-guided re-translation. The overall workflow is illustrated in Figure~\ref{fig:overall_pipeline}. Finally, we outline the experimental setup and evaluation of multiple VQA models on the proposed benchmark.


\subsection{Clinical Survey Design and Data Collection}
\label{sec:survey_method}
To understand communication practices, language barriers, and the adoption of artificial intelligence in clinical workflows, we conducted a cross-sectional survey of practicing physicians across Pakistan. The survey was designed to capture the clinical communication behavior in multilingual healthcare environments. The questionnaire was structured into four major sections: (1) demographic information, including professional role and institutional affiliation; (2) physician--patient communication practices, focusing on language usage and communication strategies; (3) use of AI tools in clinical workflows, including diagnostic support and documentation; and (4) perceptions of AI reliability, trust, and clinical responsibility. The survey instrument consisted of both multiple-choice and Likert-scale questions, along with optional open-ended responses to capture qualitative feedback. The questionnaire was developed in English and designed to be easily interpretable by clinical practitioners. Data collection was conducted using Google Forms, and participation was voluntary and anonymous. The survey was distributed through professional medical networks across multiple healthcare institutions in Pakistan, including both public and private sector hospitals. A total of 430 physicians participated in the study. No personally identifiable information was collected, ensuring participant confidentiality and compliance with ethical research standards.

\subsection{Leukemia Morphology VQA Construction}

To develop a morphology-aware WBC-VQA benchmark for leukemic and normal cells. We utilized the publicly available Leukemia patients dataset \citep{rehman2024leukemiaattri} . The dataset provides cell-level morphology annotations describing the 13 WBC classes and six clinically relevant attributes of each WBC: nuclear chromatin, nuclear shape, nucleolus characteristics, cytoplasm appearance, cytoplasmic basophilia, and cytoplasmic vacuoles, including the cell size. Although these annotations provide detailed morphological information, the dataset does not contain the textual details. Therefore, we transformed the morphology annotations into structured VQA samples to enable morphology-oriented vision-language learning and reasoning.\\
\noindent \textbf{Visual Question Answer Generation:}
We generated question answer pairs using BioMistral-7B \citep{labrak2024} following the VQA generation strategy adopted in \textsf{Uni-Hema} \citep{rehman2026uni}. For each cell image, the six morphology attributes were provided as structured input to the model. Based on these attributes, BioMistral-7B \citep{labrak2024} generated clinically relevant questions together with their corresponding answers. The generated questions focused on morphology-related characteristics commonly examined during hematological assessment, including chromatin patterns, nuclear morphology, cytoplasmic appearance, and cytoplasmic vacuoles. Depending on the available annotations, between two and six question-answer pairs were generated for each cell image.

\noindent \textbf{Prompt Design:} To generate morphology-oriented VQA pairs, morphology annotations associated with each leukemia cell were first converted into structured textual descriptions and provided as input to BioMistral-7B \citep{labrak2024}. 
The model was prompted to act as a hematology expert and generate clinically relevant question--answer pairs grounded exclusively in the provided morphology attributes. The prompt encouraged diverse question formulations, morphological reasoning, and medically consistent responses while constraining the output to the given annotations. The prompt template used for VQA generation is shown in Figure~\ref{fig:biomistral_prompt}. Using the structured prompt, BioMistral-7B generated multiple morphology-focused question--answer pairs describing key leukemia cell characteristics, including nuclear morphology, chromatin structure, cytoplasmic properties, and other clinically relevant features.

\begin{figure}[t]
\centering
\fbox{
\parbox{0.95\linewidth}{
\small
\textbf{You are creating a medical VQA dataset.}

Generate multiple question-answer pairs from this morphology.

\vspace{1mm}
\textbf{Rules:}
\begin{itemize}
\item Use provided morphology information and your medical knowledge with respect to leukemia.
\item Answers must be complete, detailed sentences.
\item Do not repeat question templates.
\item Questions may combine multiple morphology attributes.
\item Return only valid JSON.
\end{itemize}
\textbf{Morphology:}
\texttt{\{morphology\_text\}}
\vspace{2mm}
\textbf{Return format:}
\texttt{\{"q1":"...", "a1":"...", ...\}}
}
}
\caption{Prompt template used for morphology-based VQA generation with BioMistral-7B. Structured leukemia cell morphology attributes are provided as input, and the model generates multiple clinically grounded question-answer pairs in JSON format.}
\label{fig:biomistral_prompt}
\end{figure}

\noindent \textbf{Expert Validation of English VQA Samples:} The source dataset was previously validated by domain experts. Additionally, the English dataset was reviewed by a medical expert, where each question–answer pair was assessed alongside its corresponding cell image and morphology annotations for clinical relevance, clarity, and consistency. Experts verified that questions were well-formed and clinically meaningful, and that answers accurately reflected the underlying morphological attributes. Any instances of ambiguity, grammatical issues, clinical irrelevance, or annotation mismatch were corrected. Table~\ref{tab:english_validation_5col} summarizes the types of errors identified along with representative examples from the validation process.

\subsection{Urdu Dataset Construction}

The validated English question-answer pairs were translated into Urdu by using the NLLB-200 translation model \citep{nllb2022}. Translation was performed using the \texttt{eng\_Latn} to \texttt{urd\_Arab} language pair while preserving the original image-question-answer relationship.\\
\begin{table}[ht]
\centering
\caption{Examples of corrections in the Urdu dataset.}
\label{tab:urdu_term_corrections}
\resizebox{\columnwidth}{!}{%
\begin{tabular}{lll}
\hline
\textbf{English Term} & \textbf{NLLB Translation} & \textbf{Corrected Urdu} \\
\hline
cell &
{\urdufont سیل} &
{\urdufont خلیه } \\
\hline
Nucleus &
{\urdufont نیوکلیئس} &
{\urdufont مرکزہ} \\
\hline

cytoplasm
 &
{\urdufont ,سائٹوپلاسم} &
{\urdufont   مائع خلوی
} \\
\hline

Morphology &
{\urdufont مورفولوجی} &
{\urdufont ساخت} \\
\hline

Granules &
{\urdufont ,گارنٹیڈ گرانولز} &
{\urdufont  دار دانے } \\
\hline

\end{tabular}
}
\end{table}
\noindent \textbf{Translation Error Analysis:}
After translation, several inconsistencies were identified in morphology-related terminology. The Urdu dataset was therefore manually reviewed by an English-to-Urdu medical dictionary and native language speakers to detect translation errors in hematology-specific terms. It was observed that key domain concepts such as nucleus, chromatin, morphology, and other cellular attributes were sometimes translated into incorrect or clinically inappropriate Urdu terms. To resolve this issue, a domain-specific English-to-Urdu hematology dictionary was developed and applied as a post-processing step. This correction process resulted in more than five thousand terminology fixes across the dataset, and the revised translations were incorporated into the final Urdu benchmark. Table~\ref{tab:urdu_term_corrections} summarizes these corrections.
\\
\noindent \textbf{Prompt-Guided Re-translation:} After constructing the Urdu hematology dictionary, the translated dataset was reprocessed using a refined translation prompt. The translation model was explicitly instructed to preserve the standardized Urdu terminology defined in the dictionary whenever corresponding English medical terms appeared in the source text. This process reduced translation variability and improved consistency across morphology descriptions, questions, and answers, resulting in the final Urdu version of the leukemia morphology VQA dataset.

\begin{table*}[ht]
\centering
\caption{Performance comparison of VLM's on the English, Urdu, and bilingual (English+Urdu) morphology VQA benchmarks.}
\label{tab:all_results}

\resizebox{\textwidth}{!}{
\begin{tabular}{ll|ccc|ccc|ccc}
\hline
\multicolumn{11}{c}{\textbf{Zero-shot/Fine-tune and test on English}} \\
\hline
\multicolumn{2}{c}{} &
\multicolumn{3}{|c}{\textbf{LeukemiaAttri}} &
\multicolumn{3}{c}{\textbf{WBCAtt}} &
\multicolumn{3}{c}{\textbf{LeukemiaAttri + WBCAtt}} \\
\cline{3-11}
Model & Setting &
BLEU-4 & ROUGE-L & BERTScore &
BLEU-4 & ROUGE-L & BERTScore &
BLEU-4 & ROUGE-L & BERTScore \\
\hline
Qwen2-VL-2B-Instruct & Zero-shot & 0.232 & 0.409 & 0.909 & 0.090 & 0.286 & 0.878 & 0.203 & 0.384 & 0.903 \\
Qwen2-VL-2B-Instruct & Fine-tuned & 0.823 & 0.894 & 0.983 & 0.274 & 0.536 & 0.939 & 0.653 & 0.819 & 0.969 \\
InternVL2.5-2B & Fine-tuned & 0.149 & 0.261 & 0.887 & 0.052 & 0.260 & 0.886 & 0.100 & 0.261 & 0.887 \\
Uni-Hema(T5-Small) & Fine-Tuned  & 0.659 & 0.836 & -- & \textbf{0.322} & \textbf{0.557} & -- & 0.589 & 0.778 & -- \\

\hline
\multicolumn{11}{c}{\textbf{Zero-shot/Fine-tune and test on Urdu}} \\
\hline
Qwen2-VL-2B-Instruct & Zero-shot &0.049  & -- & 0.911 & 0.043 & -- & 0.910 & 0.047 & -- & 0.910 \\
Qwen2-VL-2B-Instruct & Fine-tuned & 0.112 & -- & 0.884 & 0.017 & -- & 0.842 & 0.095 & --  & 0.875 \\
\hline
\multicolumn{11}{c}{\textbf{Fine-tune on English + Urdu and test on Urdu}} \\
\hline
Qwen2-VL-2B-Instruct & Fine-tuned & 0.291 & -- & 0.953 & 0.121 & -- & 0.932 & 0.257 & -- & 0.947 \\
\hline
\multicolumn{11}{c}{\textbf{Fine-tune on English + Urdu and test on English}} \\
\hline
Qwen2-VL-2B-Instruct & Fine-tuned & \textbf{0.834} & \textbf{0.899} & \textbf{0.984} & 0.274 & 0.521 & 0.932 & \textbf{0.661} & \textbf{0.821} & \textbf{0.973} \\
\hline
\end{tabular}}

\end{table*}
\subsection{WBCAtt-VQA Bilingual Extension}
In addition to the leukemia morphology VQA dataset, we extended the validated WBCAtt-VQA dataset introduced in \textsf{Uni-Hema} \citep{rehman2026uni}. The dataset was processed using the same multilingual pipeline as the leukemia morphology dataset. Specifically, the English question–answer pairs were translated into Urdu using NLLB-200 \citep{nllb2022}, followed by translation quality assessment, dictionary-guided terminology standardization, and prompt-guided re-translation. This process resulted in a bilingual version of WBCAtt-VQA containing aligned English and Urdu annotations.
\begin{figure}[ht]
    \centering
    \includegraphics[width=\linewidth]{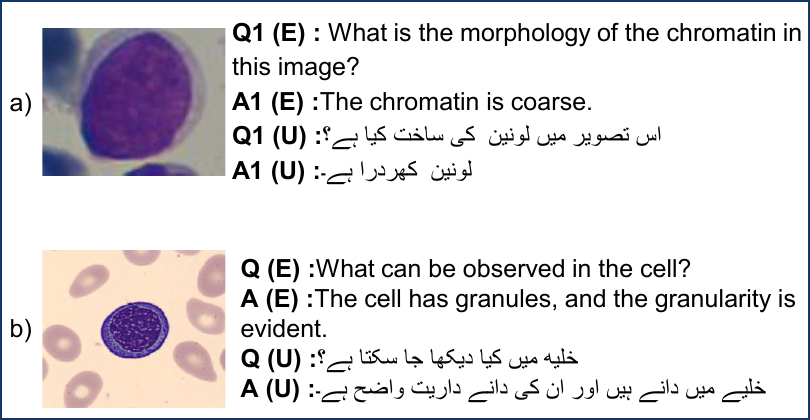}
    \caption{Examples of bilingual (English and Urdu) question--answer pairs from (a) leukemic cells and (b) normal white blood cells.}
    \label{fig:language}
\end{figure}
\subsection{Final Bilingual Benchmark}
The final benchmark consists of two clinically validated, morphology-aware, bilingual white blood cell (WBC) VQA resources for leukemic and normal cells: the newly developed leukemia morphology \citep{rehman2026uni} VQA dataset and the bilingual extension of normal morphology WBCAtt-VQA \citep{rehman2026uni}. Each sample contains a microscopic cell image together with semantically equivalent question-answer pairs in English and Urdu.
For the English benchmark, the LeukemiaAttri dataset contains 31,029 training samples and 9,424 testing samples, while the WBCAtt dataset contains 12,339 training samples and 2,468 testing samples. For the Urdu benchmark, the translated and clinically validated LeukemiaAttri dataset contains 31,029 training samples and 9,424 testing samples, whereas the Urdu WBCAtt dataset contains 12,339 training samples and 2,468 testing samples.

\subsection{Model Fine-tuning and Evaluation Setup}
\label{sec:finetuning_setup}

To demonstrate the usability of the proposed hematology VQA resource, we evaluated multiple vision-language models under zero-shot and fine-tuning settings. We evaluated two general-purpose VLMs, Qwen2-VL-2B-Instruct \citep{Qwen2VL} and InternVL2.5-2B \citep{chen2024internvl}, along with a hematology-specific unified model \citep{rehman2026uni}, to assess the dataset’s effectiveness for both general and biomedical VQA. Experiments were conducted on the WBCMor-VQA English, Urdu, and combined sets.
For fine-tuning, each sample consisted of a microscopic cell image paired with a morphology-related question and its corresponding answer. Models were trained using Low-Rank Adaptation (LoRA) \citep{che2026lora}, while full supervised fine-tuning was applied to Uni-Hema with small-T5 where applicable to improve efficiency and reduce memory overhead. The models were optimized to generate concise morphology specific responses, including nuclear chromatin pattern, nuclear shape, cytoplasmic appearance, and basophilia level. The training and test splits were kept strictly separate to avoid data leakage. Evaluation was performed on both expert-validated and non-validated test sets to analyze the impact of annotation refinement on model performance.


\section{Validation}
To validate the proposed English–Urdu hematology VQA dataset, we conduct a comprehensive evaluation using vision–language models on English and Urdu benchmarks. Strict non-overlapping train/test splits are maintained to ensure unbiased assessment and robust generalization analysis. The evaluation is performed in three settings: English-only, Urdu-only, and a combined bilingual benchmark, where the same model is directly evaluated across both languages to assess its cross-lingual generalization capability.  Model outputs are assessed using BLEU-4, ROUGE-L, and BERTScore, providing complementary lexical and semantic measures of response quality.

\subsection{Implementation Details}
\label{sec:implementation_details}

For Qwen2-VL-2B-Instruct, the maximum sequence length was set to 512 tokens, and the model was trained for three epochs with a learning rate of $5 \times 10^{-5}$, using gradient accumulation to handle limited GPU memory. For Uni-Hema, we use a pretrained T5-small model, while the rest of the network is trained from scratch using standard settings. For InternVL2.5-2B, images were resized to $448 \times 448$ pixels and the model was fine-tuned using LoRA-based supervised learning. The sequence length was set to 768 tokens, with a learning rate of $1 \times 10^{-5}$ over three epochs, along with gradient accumulation for stable optimization. During inference, greedy decoding was used with a maximum generation length of 32 tokens. Evaluation was performed using BLEU-4, ROUGE-L, Exact Match, and BERTScore F1 to assess lexical accuracy, exact match performance, and semantic similarity.

\subsection{Experiments and Results}
The performance comparison across the English, Urdu, and bilingual (English+Urdu) benchmarks is presented in Table~\ref{tab:all_results}. The results demonstrate that domain-specific fine-tuning consistently outperforms zero-shot evaluation, highlighting the effectiveness of the proposed bilingual hematology VQA dataset. Furthermore, the observed trends indicate that leveraging additional bilingual supervision enables models to generalize across both languages, suggesting that similar performance gains may be achieved when such cross-lingual resources are utilized for training other vision--language models.





\section{Conflicts of Interest}
The authors have no conflict of interest to declare.

\section{Acknowledgments}

We would like to thank Saadia Malik, Zahra Malik, and Fatima Athar Khan for their continuous support, encouragement, and valuable contributions throughout this research. Their assistance is sincerely appreciated and played an important role in the completion of this work.

\bibliography{sample}


\end{document}


\section*{Supplementary Material}

This supplementary material provides additional details for the proposed
English--Urdu hematopathology visual question answering dataset. It includes
Urdu terminology refinement examples, dataset statistics, physician survey
details, survey questionnaire sections, and additional survey result figures.

\section{Urdu Medical Terminology Refinement}

Table~\ref{tab:urdu_dictionary} presents examples of terminology corrections
identified during Urdu dataset validation. These corrections were applied to
improve linguistic consistency and clinical accuracy in the translated Urdu
question-answer pairs.

\renewcommand{\arraystretch}{1.35}

\begin{longtable}{|p{3cm}|p{6cm}|p{5cm}|}
\caption{Examples of Urdu terminology refinement. The table shows selected
medical and morphology-related English terms, their machine-generated Urdu
translations, and the corrected Urdu terms used after validation.}
\label{tab:urdu_dictionary} \\
\hline
\rowcolor{gray!20}
\textbf{English Term} & \textbf{Model Urdu Translation} & \textbf{Corrected Urdu Term} \\
\hline
\endfirsthead

\hline
\rowcolor{gray!20}
\textbf{English Term} & \textbf{Model Urdu Translation} & \textbf{Corrected Urdu Term} \\
\hline
\endhead

Acute & {\urdufont تیز رفتار} & {\urdufont شدید} \\
\hline
Appearance & {\urdufont شکل} & {\urdufont ظاہری شکل} \\
\hline
Band & {\urdufont بینڈ} & {\urdufont پٹی نما} \\
\hline
Basophil & {\urdufont باسوفیل} & {\urdufont بیسوفل} \\
\hline
Cell & {\urdufont سیل} & {\urdufont خلیہ} \\
\hline
Chromatin & {\urdufont کروماتین، کروماٹین، کرومٹین} & {\urdufont لونین} \\
\hline
Coarse & {\urdufont گہرا، بھاری} & {\urdufont کھردرا} \\
\hline
Cytological & {\urdufont سائٹولوجیکل} & {\urdufont خلویاتی} \\
\hline
Cytoplasm & {\urdufont سائٹوپلاسما، سائٹوپلاس، مادہما، سیٹوپلاسما، مادہم} & {\urdufont خلوی مادہ} \\
\hline
Cytoplasmic & {\urdufont سائٹوپلاسٹک، سائٹوپلاسمیک} & {\urdufont خلیہ مایہ} \\
\hline
Degree & {\urdufont سطح، ڈگری} & {\urdufont مقدار} \\
\hline
Granular & {\urdufont گارنٹیڈ، گارنٹی} & {\urdufont دانے دار} \\
\hline
Hypersegmented & {\urdufont ہائپر سیگمنٹڈ} & {\urdufont کثیر فصوصی} \\
\hline
Indented & {\urdufont چھلانگ لگا} & {\urdufont نشیب دار} \\
\hline
Lobated & {\urdufont لوبیٹڈ} & {\urdufont لوب بننا} \\
\hline
Lobed & {\urdufont لوب} & {\urdufont گوشہ دار} \\
\hline
Moderate & {\urdufont اعتدال پسند} & {\urdufont معتدل} \\
\hline
Morphology & {\urdufont مورفولوجی} & {\urdufont ساخت} \\
\hline
Multilobed & {\urdufont کثیر لابی، کثیر ذرو، کثیر خلیات} & {\urdufont کثیرلختہ} \\
\hline
Myeloblast & {\urdufont مائیلو بلاسٹ} & {\urdufont نخاعی خلیہ} \\
\hline
Myelocytic & {\urdufont مائیلوسیٹک} & {\urdufont نخاعی خلیاتی} \\
\hline
Myeloid & {\urdufont مائیلوئڈ} & {\urdufont ہڈی کا گودا} \\
\hline
Neutrophilic & {\urdufont نیوٹروفیل} & {\urdufont عدلیاتی} \\
\hline
Nuclear & {\urdufont نیوکلیئر} & {\urdufont جوہری} \\
\hline
Nucleus & {\urdufont نوکیولس، نوکیلیوس، نیوکلئس، نوکلہ، جوہری} & {\urdufont مرکزہ} \\
\hline
Oval & {\urdufont اوال} & {\urdufont بیضوی} \\
\hline
Pattern & {\urdufont پیٹرن} & {\urdufont خاکہ} \\
\hline
Sarcoma & {\urdufont سارکومہ} & {\urdufont لحمومہ} \\
\hline
Size & {\urdufont سائز} & {\urdufont پیمائش} \\
\hline
Thin rim & {\urdufont پتلے کنارے} & {\urdufont باریک جھلی} \\
\hline
Trilobed & {\urdufont ٹرپلوبڈ} & {\urdufont سَہ لختہ} \\
\hline
Vacuoles & {\urdufont مایہلونین، ویکیولز} & {\urdufont خلا} \\
\hline

\end{longtable}

\section{Dataset Details}

This section summarizes the final bilingual dataset statistics. Each image-level
sample contains one English question-answer pair and one corresponding Urdu
question-answer pair. Therefore, the total number of QA pairs represents the
combined English and Urdu annotations.

\subsection{Leukemia Dataset}

\begin{table}[H]
\centering
\caption{Statistics of the leukemia VQA dataset. The dataset contains
cell-level morphology-based English and Urdu question-answer pairs generated
from leukemia cell images.}
\label{tab:leukemia_dataset}
\resizebox{\linewidth}{!}{
\begin{tabular}{lcccccc}
\toprule
\textbf{Split} & \textbf{No. of Cells} & \textbf{English Q} & \textbf{English A} & \textbf{Urdu Q} & \textbf{Urdu A} & \textbf{Total QA} \\
\midrule
Train & 31,029 & 31,029 & 31,029 & 31,029 & 31,029 & 62,058 \\
Test  & 9,424  & 9,424  & 9,424  & 9,424  & 9,424  & 18,848 \\
\bottomrule
\end{tabular}
}
\end{table}

\subsection{WBC-Att Dataset}

\begin{table}[H]
\centering
\caption{Statistics of the WBC-Att VQA dataset. This subset contains
morphology-based question-answer pairs for normal white blood cell images.}
\label{tab:wbcatt_dataset}
\resizebox{\linewidth}{!}{
\begin{tabular}{lcccccc}
\toprule
\textbf{Split} & \textbf{No. of Cells} & \textbf{English Q} & \textbf{English A} & \textbf{Urdu Q} & \textbf{Urdu A} & \textbf{Total QA} \\
\midrule
Train & 12,339 & 12,339 & 12,339 & 12,339 & 12,339 & 24,678 \\
Test  & 2,468  & 2,468  & 2,468  & 2,468  & 2,468  & 4,936 \\
\bottomrule
\end{tabular}
}
\end{table}

\subsection{Combined Leukemia and WBC-Att Dataset}

\begin{table}[H]
\centering
\caption{Statistics of the combined bilingual hematopathology VQA dataset.
The merged dataset includes both leukemia and normal white blood cell samples
for multilingual VQA training and evaluation.}
\label{tab:combined_dataset}
\resizebox{\linewidth}{!}{
\begin{tabular}{lcccccc}
\toprule
\textbf{Split} & \textbf{No. of Cells} & \textbf{English Q} & \textbf{English A} & \textbf{Urdu Q} & \textbf{Urdu A} & \textbf{Total QA} \\
\midrule
Train & 43,468 & 43,468 & 43,468 & 43,468 & 43,468 & 86,936 \\
Test  & 11,892 & 11,892 & 11,892 & 11,892 & 11,892 & 23,784 \\
\midrule
Total & 55,360 & 55,360 & 55,360 & 55,360 & 55,360 & 110,720 \\
\bottomrule
\end{tabular}
}
\end{table}

\noindent\textbf{Total Dataset Size:} 110,720 bilingual QA annotations.

\section{Physician Survey Details}

A cross-sectional physician survey was conducted to study clinical
communication practices, language preferences, patient misunderstanding, and
the perceived role of AI-assisted medical systems in Pakistan.

\begin{table}[H]
\centering
\caption{Summary of the physician survey timeline and access details.}
\label{tab:survey_details}
\begin{tabular}{ll}
\toprule
\textbf{Item} & \textbf{Details} \\
\midrule
Start Date & January 28, 2026 \\
End Date & June 13, 2026 \\
Survey Link & \url{https://docs.google.com/forms/d/11bqfPbY14vpb25DP2QM3CZDJIDr5jFHsHVpPdF6e0g8/viewform} \\
\bottomrule
\end{tabular}
\end{table}

\section{Survey Questionnaire}

\begin{figure}[H]
    \centering
    \includegraphics[width=0.82\linewidth]{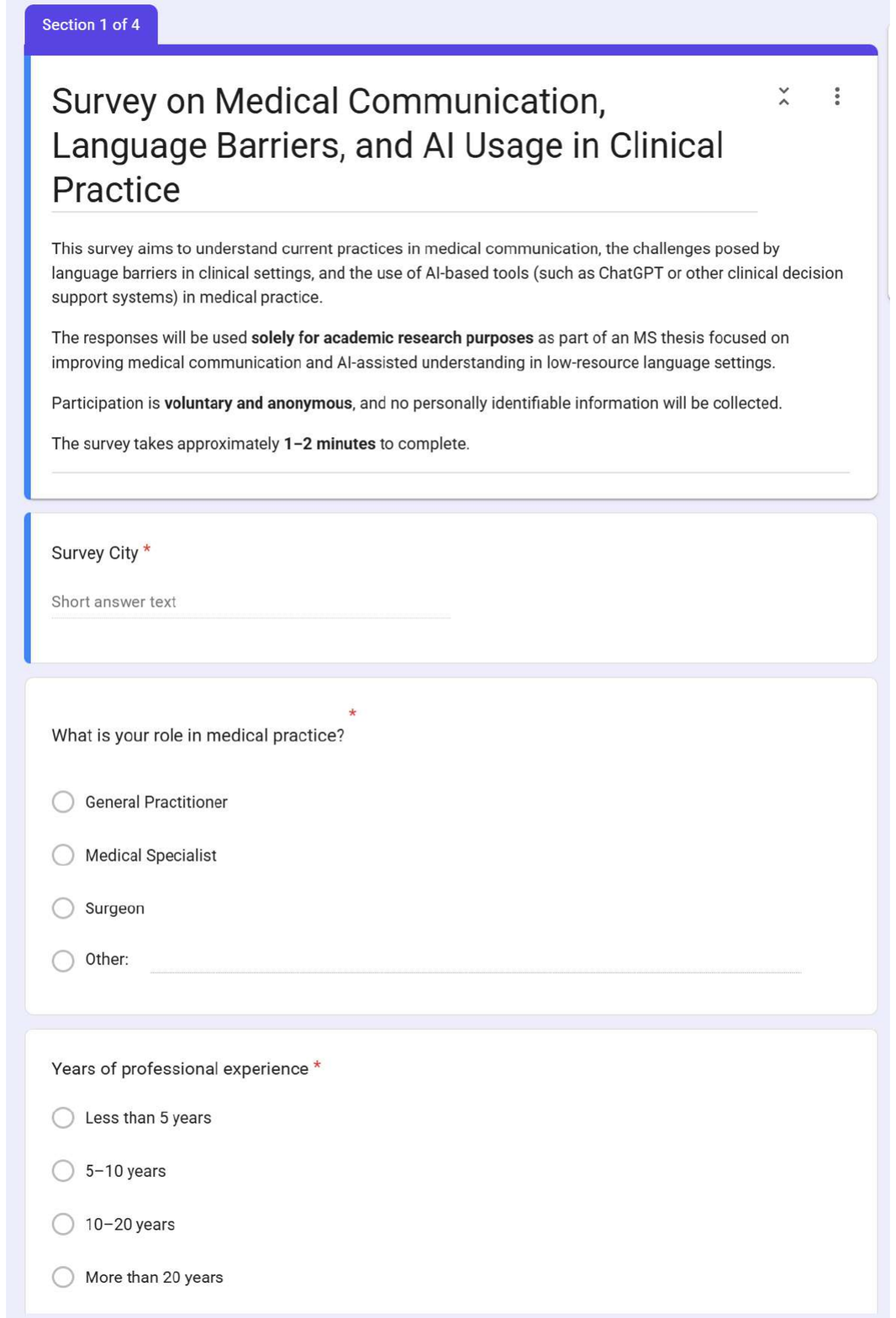}
    \caption{Section 1 of the physician survey questionnaire. This section
    collected participant demographics, including age group, gender, clinical
    experience, medical specialty, and healthcare sector.}
    \label{fig:survey_section_1}
\end{figure}

\begin{figure}[H]
    \centering
    \includegraphics[width=0.82\linewidth]{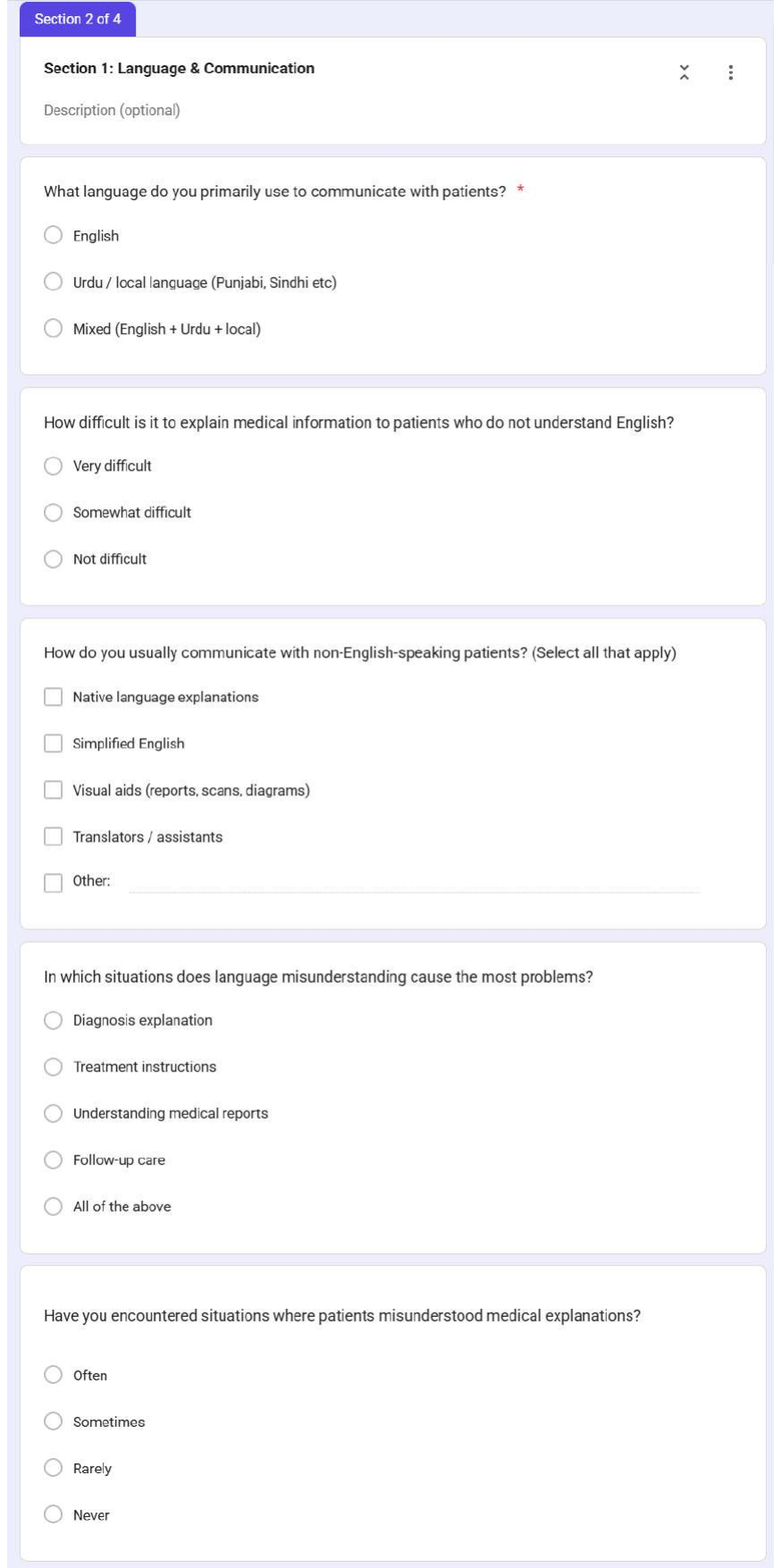}
    \caption{Section 2 of the questionnaire focusing on physician-patient
    communication practices, preferred language of communication, and the
    frequency of communication barriers in routine clinical settings.}
    \label{fig:survey_section_2}
\end{figure}

\begin{figure}[H]
    \centering
    \includegraphics[width=0.82\linewidth]{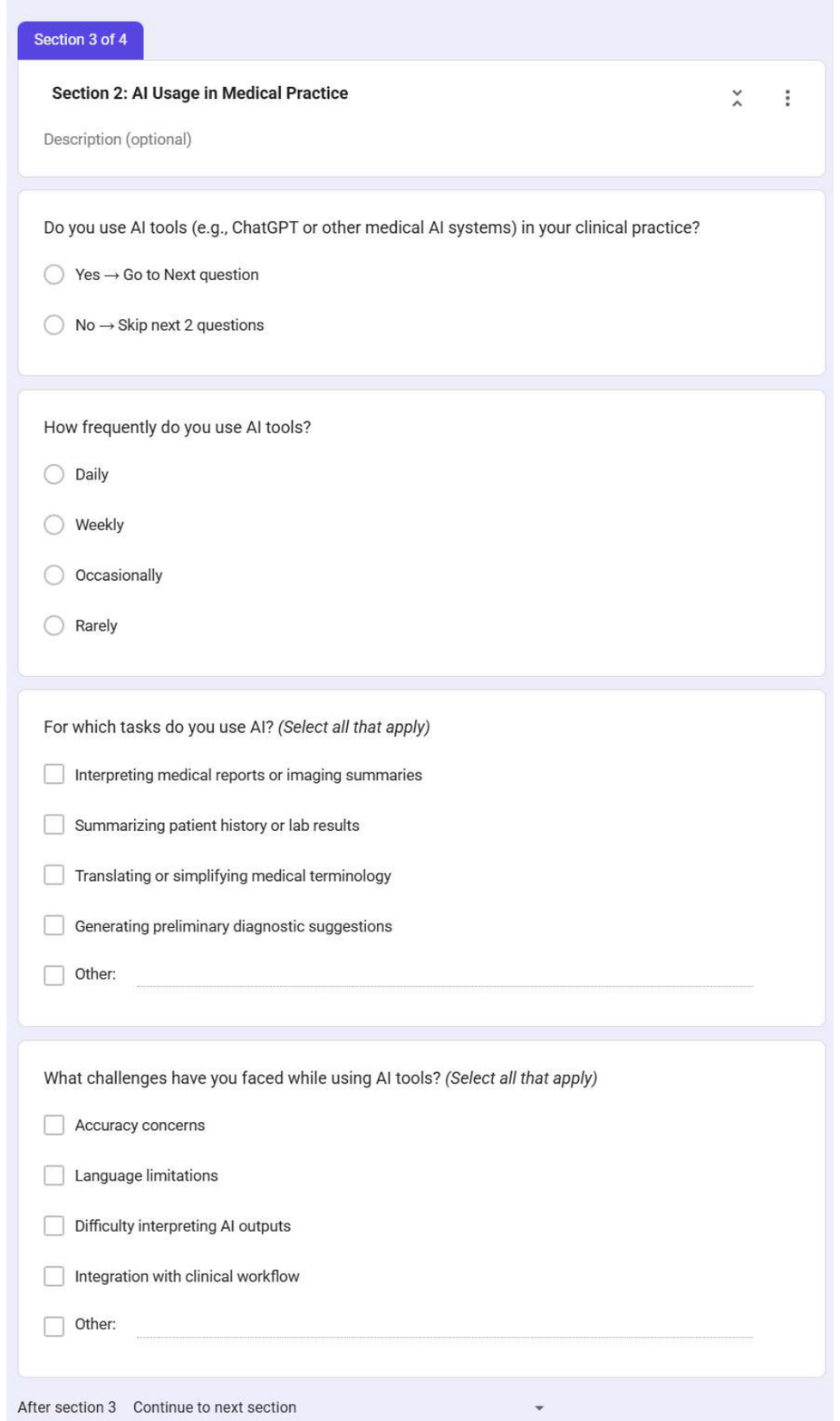}
    \caption{Section 3 of the questionnaire investigating physician experience
    with AI-assisted medical systems, including usage patterns, perceived
    benefits, and concerns related to clinical deployment.}
    \label{fig:survey_section_3}
\end{figure}

\begin{figure}[H]
    \centering
    \includegraphics[width=0.82\linewidth]{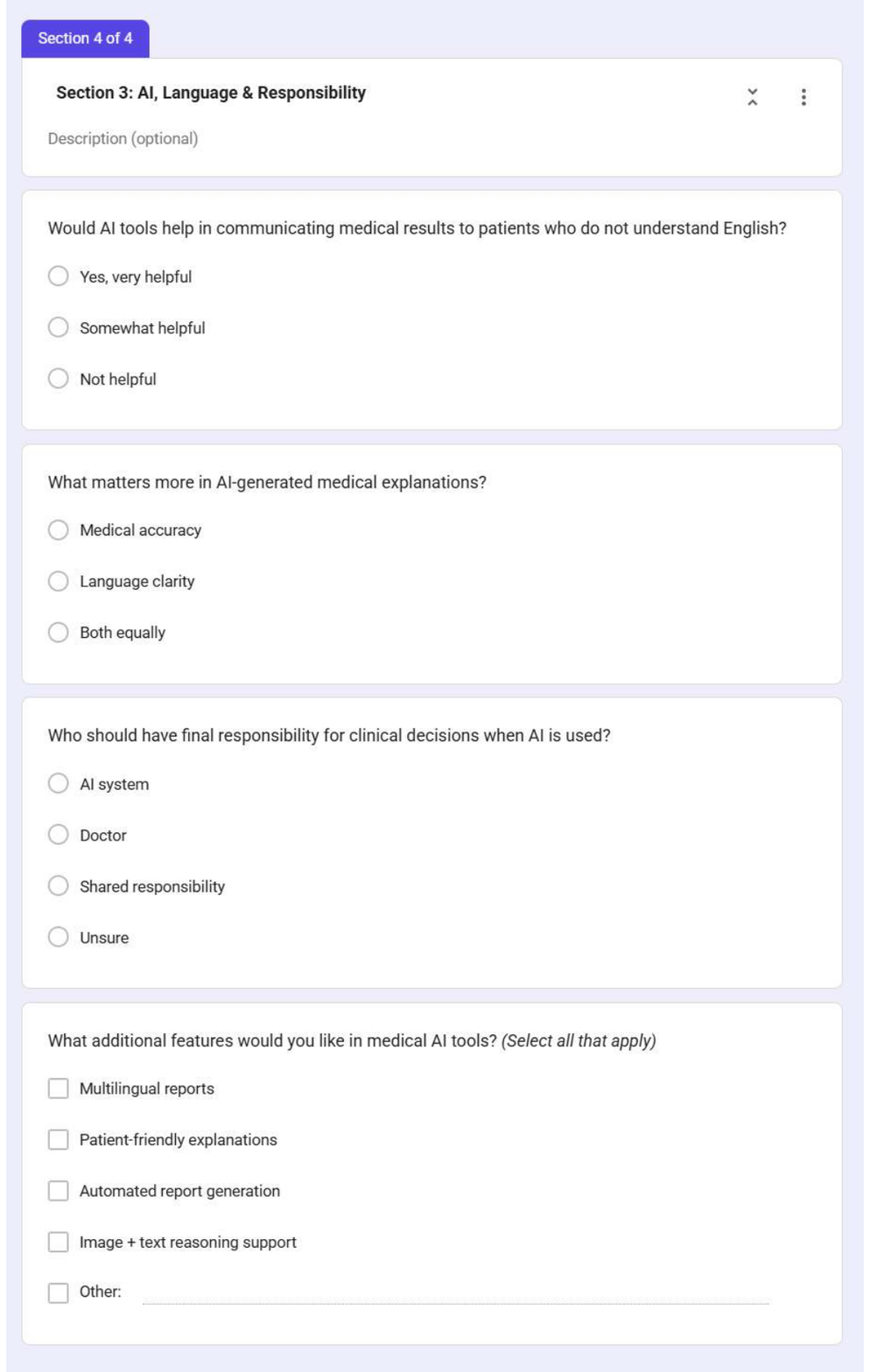}
    \caption{Section 4 of the questionnaire examining attitudes toward
    multilingual medical AI systems and the perceived importance of Urdu
    language support for improving accessibility and patient understanding.}
    \label{fig:survey_section_4}
\end{figure}

Figures~\ref{fig:survey_section_1}--\ref{fig:survey_section_4} show the
main sections of the physician survey questionnaire.

\section{Survey Result Figures}

This section presents additional survey result visualizations that support the
motivation for developing bilingual medical AI resources.

\section{Additional Survey Result Figures}

\begin{figure}[H]
    \centering
    \includegraphics[width=0.72\linewidth]{figures/Professional Experience.pdf}
    \caption{Distribution of physicians by professional experience. More than half of the respondents had less than five years of clinical experience, while the remaining participants represented mid-career and senior clinical experience groups.}
    \label{fig:professional_experience}
\end{figure}

\begin{figure}[H]
    \centering
    \includegraphics[width=0.72\linewidth]{figures/Use of AI Tools.pdf}
    \caption{Use of AI tools among surveyed physicians. A majority of respondents reported using AI-assisted tools in some form, indicating increasing clinical familiarity with AI-based support systems.}
    \label{fig:use_ai_tools}
\end{figure}

\begin{figure}[H]
    \centering
    \includegraphics[width=0.78\linewidth]{figures/AI Tasks Used by Physicians.pdf}
    \caption{Clinical tasks for which physicians reported using AI tools. Diagnostic suggestions and medical terminology translation were the most common uses, followed by report or imaging interpretation and patient history or laboratory summary generation.}
    \label{fig:ai_tasks}
\end{figure}

\begin{figure}[H]
    \centering
    \includegraphics[width=0.72\linewidth]{figures/AI Helps Communicate Medical Results.pdf}
    \caption{Physician perception of whether AI tools help communicate medical results to patients. Most respondents considered AI somewhat helpful or very helpful, suggesting potential value for patient-facing explanation support.}
    \label{fig:ai_communication_help}
\end{figure}

\begin{figure}[H]
    \centering
    \includegraphics[width=0.78\linewidth]{figures/Situations Where Language Misunderstanding Causes Problems.pdf}
    \caption{Clinical situations where language misunderstanding causes communication problems. Diagnosis explanation, treatment instructions, medical report understanding, and follow-up care were identified as key areas affected by language barriers.}
    \label{fig:language_misunderstanding_situations}
\end{figure}

\section{Qualitative Model Prediction Results}

This section presents qualitative examples from the model prediction files.
The examples include correct predictions, clinically related paraphrases, and
failure cases from both leukemia and normal WBC test samples.

\begin{table*}[t]
\centering
\caption{
Qualitative examples of model predictions on leukemia and normal WBC VQA
samples. Correct cases show exact or near-exact agreement with the reference
answer. Partially correct cases preserve some clinically relevant information
but miss specific morphology details. Failure cases show weak alignment, often
because the model produces a generic explanation instead of answering the
specific morphology question.
}
\label{tab:qualitative_model_predictions}
\resizebox{\textwidth}{!}{
\begin{tabular}{p{2.2cm}p{3.3cm}p{4.0cm}p{4.0cm}p{4.0cm}p{4.6cm}}
\toprule
\textbf{Dataset} & \textbf{Case Type} & \textbf{Question} & \textbf{Reference Answer} & \textbf{Model Prediction} & \textbf{Analysis} \\
\midrule

Leukemia &
Correct &
What is the morphology of the nuclear shape in this cell? &
The nuclear shape in this cell is irregular. &
The nuclear shape in this cell is irregular. &
The model exactly matches the reference answer and correctly identifies the irregular nuclear shape. \\

\midrule

Leukemia &
Correct &
What is the degree of cytoplasmic basophilia in this cell? &
The degree of cytoplasmic basophilia in this cell is moderate. &
The degree of cytoplasmic basophilia in this cell is moderate. &
The prediction is fully aligned with the ground truth and correctly captures the cytoplasmic basophilia degree. \\

\midrule

Leukemia &
Partial &
What is the nuclear chromatin morphology of the leukemic cells? &
The nuclear chromatin of the leukemic cells is coarse. &
The nuclear chromatin morphology of the leukemic cells is hyperchromatic and clumped. &
The prediction is clinically related because hyperchromatic and clumped chromatin can describe abnormal chromatin appearance, but it does not exactly match the reference term ``coarse.'' \\

\midrule

Leukemia &
Failure &
What are the cytoplasmic vacuoles? &
The cytoplasmic vacuoles are absent. &
Cytoplasmic vacuoles are intracellular structures that contain fluid and are typically found in cells that are undergoing cell division. &
The model gives a general definition instead of answering the image-specific question. It fails to identify that vacuoles are absent in the reference annotation. \\

\midrule

WBC &
Correct &
Q: What is depicted in this image? &
This image depicts a neutrophil, which is a type of white blood cell. &
This image depicts a neutrophil, which is a type of white blood cell. &
The model exactly matches the reference answer and correctly identifies the WBC type. \\

\midrule

WBC &
Correct &
Q: What is the texture of the cell? &
The cell has a granular texture. &
The cell has a granular texture. &
The prediction is fully correct and captures the expected morphology descriptor. \\

\midrule

WBC &
Partial &
Q: What is the type of white blood cell shown in the image? &
The type of white blood cell shown in the image is a neutrophil. &
The cell in the image is a neutrophil. &
The model uses shorter wording but preserves the same clinical meaning by correctly identifying the cell as a neutrophil. \\

\midrule

WBC &
Failure &
Q: What is the number of the granules in the cell? &
The granules are not very numerous throughout the cell. &
4 &
The model gives a numeric answer instead of describing the granule distribution. This shows a failure in understanding the expected descriptive morphology response. \\

\bottomrule
\end{tabular}
}
\end{table*}

```latex
\section{Qualitative Urdu VQA Results}

This section presents qualitative examples of Urdu visual question answering
predictions generated by the Qwen2-VL model. The examples illustrate both
successful morphology recognition and challenging failure cases observed
during evaluation.

\begin{table*}[t]
\centering
\caption{
Successful Urdu VQA predictions generated by the Qwen2-VL model. The model
correctly identifies clinically relevant morphology attributes including
chromatin structure, nuclear shape, and cytoplasmic abundance. The examples
demonstrate strong semantic alignment between the generated answers and the
expert-validated reference annotations.
}
\label{tab:urdu_success}
\renewcommand{\arraystretch}{1.2}
\resizebox{\textwidth}{!}{
\begin{tabular}{p{4.5cm}p{4.2cm}p{4.2cm}p{5.2cm}}
\toprule
\textbf{Question} &
\textbf{Ground Truth} &
\textbf{Prediction} &
\textbf{Analysis} \\
\midrule

{\urdufont اس خلیہ میں کروماتین کی ساخت کیا ہے؟}
&
{\urdufont اس خلیہ میں کروماتین کی ساخت کھردری ہے۔}
&
{\urdufont اس خلیہ میں کروماتین کی ساخت کھردری ہے۔}
&
The prediction exactly matches the reference answer and correctly identifies the chromatin morphology. \\

\midrule

{\urdufont فراہم کردہ تصویر میں جوہری شکل کیا ہے؟}
&
{\urdufont فراہم کردہ تصویر میں جوہری شکل غیر منظم ہے۔}
&
{\urdufont فراہم کردہ تصویر میں جوہری شکل غیر منظم ہے۔}
&
The model accurately recognizes the nuclear morphology and produces an exact answer. \\

\midrule

{\urdufont فراہم کردہ تصویر میں سائٹوپلاسما کی مقدار کتنی ہے؟}
&
{\urdufont فراہم کردہ تصویر میں سائٹوپلاسما کی مقدار بہت زیادہ ہے۔}
&
{\urdufont فراہم کردہ تصویر میں سائٹوپلاسما کی مقدار بہت زیادہ ہے۔}
&
The prediction correctly captures the abundance of cytoplasm and fully aligns with the annotation. \\

\midrule

{\urdufont لیوکیمیا خلیات میں سائٹوپلاسما کی مقدار کتنی ہے؟}
&
{\urdufont لیوکیمیا خلیات میں سائٹوپلاسما کی مقدار بہت زیادہ ہے۔}
&
{\urdufont لیوکیمیا خلیات میں سائٹوپلاسما کی مقدار بہت زیادہ ہے۔}
&
The generated answer preserves the complete clinical meaning and matches the reference annotation. \\

\bottomrule
\end{tabular}
}
\end{table*}

\begin{table*}[t]
\centering
\caption{
Failure cases from the Urdu VQA benchmark. Although the model often generates
fluent Urdu text, it occasionally confuses cell identity, produces uncertain
responses, or fails to recover the correct morphology attribute from the
microscopic image.
}
\label{tab:urdu_failure}
\renewcommand{\arraystretch}{1.2}
\resizebox{\textwidth}{!}{
\begin{tabular}{p{4.5cm}p{4.2cm}p{4.2cm}p{5.2cm}}
\toprule
\textbf{Question} &
\textbf{Ground Truth} &
\textbf{Prediction} &
\textbf{Analysis} \\
\midrule

{\urdufont س: تصویر میں سفید خون کے خلیات کی قسم کیا ہے؟}
&
{\urdufont تصویر میں سفید خون کے خلیات کی قسم ایک نیوٹروفیل ہے۔}
&
{\urdufont تصویر میں سفید خون کے خلیات کی قسم کم ہے۔}
&
The model confuses cell identity with a quantitative descriptor and fails to identify the correct white blood cell type. \\

\midrule

{\urdufont س: اس تصویر میں سفید خون کا خلیہ کیا ہے؟}
&
{\urdufont اس تصویر میں سفید خون کا خلیہ ایک نیوٹروفیل ہے۔}
&
{\urdufont اس تصویر میں سفید خون کا خلیہ غیر واضح ہے۔}
&
The prediction does not correctly recognize the cell category despite the presence of morphology information in the image. \\

\midrule

{\urdufont س: اس خلیہ مایہ کا رنگ کیا ہے؟}
&
{\urdufont خلیہ مایہ ہلکے نیلے رنگ کا ہے۔}
&
{\urdufont خلیہ مایہ کا پتہ نہیں چلتا۔}
&
The model fails to identify the cytoplasmic color and instead generates an uncertain response. \\

\bottomrule
\end{tabular}
}
\end{table*}
\section{Additional Notes}

The supplementary figures and tables provide extended evidence for the need
for clinically validated bilingual hematopathology VQA resources. The dataset
statistics demonstrate the scale of the English--Urdu benchmark, while the
survey results support the importance of language accessibility in medical AI
systems designed for local clinical environments.